%% file: main.tex
\documentclass[runningheads]{llncs}
\usepackage{graphicx}
\usepackage{color} 
\usepackage{comment} 
\usepackage{amsmath}
\usepackage{algorithm}
\usepackage[noend]{algorithmic}
\usepackage[misc]{ifsym}
\usepackage{wrapfig}
\usepackage{multirow}

\begin{document}

\title{
% How to Utilize Failure Demo Data?\\
% Effective Data Selection for Imitation Learning Using Distribution Differences in Attention Mechanism
How to Utilize Failure Demo Data?:
Effective Data Selection for Imitation Learning
Using Distribution Differences in Attention Mechanism
}

\titlerunning{
Effective Failure Data Selection for Imitation Learning
}

%Imitation Learning Methods Considering Differences Between Success and Failure Data in Robot Tasks
% Effective failure data selection for imitation learning using distribution differences


\author{
Kana Miyamoto\inst{1}
Kanata Suzuki\inst{2} \and 
Tetsuya Ogata\inst{1}(\Letter)
}
\authorrunning{
K. Miyamoto, K. Suzuki and T. Ogata
}

\institute{
$^{1}$ Faculty of Science and Engineering, Waseda University, Tokyo, Japan \\
\email{ogata@waseda.jp} \\
$^{2}$ Physical AI Laboratory, Fujitsu Limited, Kanagawa, Japan \\
\email{suzuki.kanata@jp.fujitsu.com}
}

\maketitle

\input{sections/0_abst}

\input{sections/1_intro}
\input{sections/2_related}
\input{sections/3_method}
\input{sections/4_experiments}
\input{sections/5_result}
\input{sections/6_conclusion}

\section*{Acknowledgment}
This work was supported by JST PRESTO, Japan, Grant Number JPMJPR24T4.

\bibliographystyle{unsrt}
\bibliography{ref}

\end{document}

%% file: sections/0_abst.tex
\begin{abstract}
Imitation learning for robotic tasks has relied primarily on policies trained only on successful demonstrations, although failures are unavoidable during human data collection.
Many existing approaches for exploiting failure data require additional data processing or iterative policy updates through autonomous rollouts, making it difficult to directly and stably utilize failure data accumulated during data collection.
In this work, we propose a method that learns latent representations of success–failure discrepancies and incorporates them into the attention mechanism.
During inference, an appropriate latent mode is selected from the initial observation to improve action stability.
Furthermore, we introduce a post-training metric that quantifies the attention discrepancy between each failure sample and successful demonstrations to select failure data.
Simulation results show that the proposed method improves task success rates when trained with failure data and that the proposed metric identifies failure samples that are beneficial for learning when combined with successful demonstrations.
These results suggest that the proposed method can support more efficient use of collected demonstrations in robotic data collection pipelines.

\keywords{Imitation learning \and Robotics \and Failure data}
\end{abstract}

%% file: sections/1_intro.tex
\section{Introduction}

Imitation learning~\cite{hussein2017imitation,zare2024survey} is a promising framework for robots operating in diverse environments such as daily living spaces, as it enables policies to be learned directly from human demonstrations.
However, its performance largely depends on the quantity and quality of training data, and collecting sufficient data is costly~\cite{agibot}.
In robotic tasks, many imitation learning models are commonly trained from successful demonstrations~\cite{ACT,diffusionpolicy}, which makes the collection of sufficient successful data a substantial burden. 
% cumbersome.
Therefore, methods that improve performance by effectively utilizing previously collected data are required.

During manual data collection, task failures due to operational mistakes or environmental factors are unavoidable, and real-world datasets often contain both successful and failed trajectories~\cite{bu2024aligning}.
The exploitation of such naturally accumulated failure data can enable more efficient learning.
Accordingly, recent studies have explored failure detection, failure avoidance, and learning methods that incorporate failure experience~\cite{fortress,inami2025motion,Fail2Progress,pi06,ssdf,irlf,IROS2021,aha,wang2024imitation}.
However, many of these methods rely on post hoc data editing or additional data collection and retraining through autonomous rollouts, which pose challenges with respect to their operational cost and scalability.
Therefore, a framework is needed to stably utilize failure data already included in collected datasets without the need for additional data collection or large-scale data correction.

Nevertheless, adding failure data to training may destabilize learning because successful and failed trajectories follow different action distributions.
Thus, exploiting failure data accumulated during data collection is difficult without careful selection.
Prior studies have reported that policy learning can be improved by removing inaccurate demonstrations in advance, or by selectively using informative segments from failed trajectories~\cite{ssdf,CoRL2023}.
% Therefore, to effectively leverage failures, learning representations that capture the differences between success and failure, and providing a mechanism for switching to appropriate behavior as needed, are necessary.
Therefore, effectively leveraging failures requires representations that capture success--failure differences and a mechanism for switching to appropriate behavior depending on the situation.

In this study, we incorporate into the loss a training signal based on the success/failure label of each sample to increase the difference between the corresponding attention distributions, thereby encouraging latent representations that better distinguish success from failure.
These latent representations are reflected in action generation through the attention mechanism, and by selecting a latent mode corresponding to successful data based on the initial observation, the method aims to generate stable actions guided toward successful behavior.
% Furthermore, the proposed method uses post-training differences in attention distributions as an analysis metric to identify failure data that contribute effectively compared with successful data, improving the efficiency of failure data utilization.
Furthermore, the proposed method uses post-training differences in attention distributions as a metric to identify failure samples that contribute effectively when combined with successful demonstrations, thereby enhancing the efficiency of their utilization.
Experiments performed on a simulated cube-lifting task demonstrate the effectiveness of the proposed method.
We evaluate both the effect of incorporating failure data during training and the influence of different failure data selection strategies on task success.
The primary contributions of this study are as follows:
\begin{itemize}
    \item We introduce latent representations that capture success--failure distributional differences into the attention mechanism to stabilize imitation learning with failure data.
    \item We propose an attention-discrepancy-based failure data selection method and show that selecting failures that complement successful data improves action performance.
\end{itemize}

%% file: sections/2_related.tex
\section{Related Works}

We categorize prior studies on handling failure data in robotic manipulation tasks into three groups: using failures in evaluative learning; improving the experience distribution through human intervention and failure correction; and incorporating information from imperfect demonstrations containing failures into learning.

First, some approaches incorporate failures into evaluative learning by treating them as negative examples or low-return trials and learning an evaluation model that estimates the success probability or action value.
Levine et al.~\cite{levine2018} trained a CNN to predict grasp success probability from monocular images and used these predictions for visual-servo-based grasping.
In this framework, success/failure labels serve as supervision for the success predictor.
Recent vision-language-action (VLA) models have also explored policy improvement from execution data collected during autonomous rollouts.
$\pi^{*}_{0.6}$~\cite{pi06} integrates demonstrations, autonomous rollouts, and expert interventions, and iteratively retrains the policy using values estimated from success/failure feedback.
While such approaches can exploit failures as learning signals, they often require reward design, success/failure feedback, additional data collection, and iterative retraining.

Next, some methods do not use failures directly as learning signals, but instead improve the experience distribution by correcting or reconstructing the training data in advance, such as by replacing failure-prone situations with successful experiences.
Inami et al.~\cite{inami2025motion} proposed a method that performs post-editing while replaying collected motion data and modifies parts of the trajectory toward successful behavior.
In $\pi^{*}_{0.6}$~\cite{pi06}, human intervention data are also collected during autonomous robot execution, where an operator guides the robot toward successful trajectories through teleoperation.
Although such human-in-the-loop correction can locally address situations prone to failure, it requires substantial human effort for data editing, monitoring, and intervention during robot operation, and may depend heavily on the task and environment.

Furthermore, several studies incorporate information from imperfect demonstrations containing failures into learning.
Shiarlis et al.~\cite{irlf} showed that the use of failed demonstrations with successful ones can reduce ambiguity in reward estimation in inverse reinforcement learning.
Hertel et al.~\cite{IROS2021} proposed a method for reproducing behavior from both successful and failed examples by optimizing for both attraction toward successful demonstrations and repulsion from failed ones.
Wu et al.~\cite{ssdf} presented a framework that performs self-supervised selection of useful segments from failed trajectories and adds them to the training data, rather than using failed trajectories directly.
These studies are closely related to our work because they exploit information in failure data.

In this study, we aim to directly incorporate failure data into offline imitation learning without relying on human intervention or iterative evaluation and retraining after deployment. In particular, our method is characterized by learning the distributional difference between success and failure as a latent representation through differences in attention distributions that are based on success/failure labels.
This representation is then used both for action generation at inference time and for failure data selection. 
If failure data can be effectively utilized in offline imitation learning, it may also provide a potential direction for the further scaling of recent VLA models.

%% file: sections/3_method.tex
\section{Method}

\begin{wrapfigure}{r}{0.5\columnwidth}
    \centering
    \vspace{-1.0em}
    \includegraphics[width=0.46\columnwidth, trim=0 0 120 0, clip]{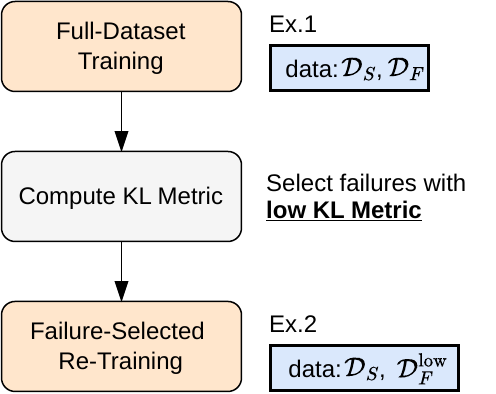}
    \caption{Training pipeline of the proposed method.}
    \label{fig:model_pipeline}
    \vspace{-1.0em}
\end{wrapfigure}

In this study, we assume that the collected demonstration dataset consists of a success subset $\mathcal{D}_S$ and a failure subset $\mathcal{D}_F$, where each demonstration has a success/failure label.
The proposed framework consists of two training processes (Fig.~\ref{fig:model_pipeline}).
In the first process, an imitation learning model is trained on all successful and failed demonstrations, $\{\mathcal{D}_S, \mathcal{D}_F\}$ (Full-Dataset Training).
% An imitation learning model is trained using all successful and failed demonstrations, $\{D_S, D_F\}$ (Full-Dataset Training). We employ a transformer encoder-decoder [16] that incorporates parametric bias (PB) [17, 18], a learnable latent representation, into its multi-head attention module.
We adopt a transformer encoder\nobreakdash-decoder model~\cite{ACT} and introduce parametric bias (PB)~\cite{PB,suzuki-iros24}, a learnable latent representation, into the multi-head attention module of the transformer decoder.
PB provides a compact way to condition the model's behavior without substantially modifying the main network architecture.
PB is incorporated into decoder self-attention, and Kullback--Leibler (KL) regularization is introduced to encourage PB to capture differences between the attention distributions of successful and failed demonstrations.
In the second process, after initial training, the attention-based KL metric with respect to demonstrations is computed for each failure sample and used as an evaluation metric. 
Using this metric, we select failure samples with small distributional discrepancy from successful demonstrations and reconstruct the training dataset as $\{\mathcal{D}_S, \mathcal{D}_F^{\mathrm{low}}\}$.
The model is then retrained on the reconstructed dataset (Failure-Selected Re-Training).
The following subsections describe PB and KL regularization (Sec.~\ref{sec:pb_kl}), KL-based failure data selection and retraining (Sec.~\ref{sec:kl_select}), and PB selection at inference time (Sec.~\ref{sec:pb_select}).

\begin{figure}[tb]
    \centering
    \includegraphics[width=\columnwidth]{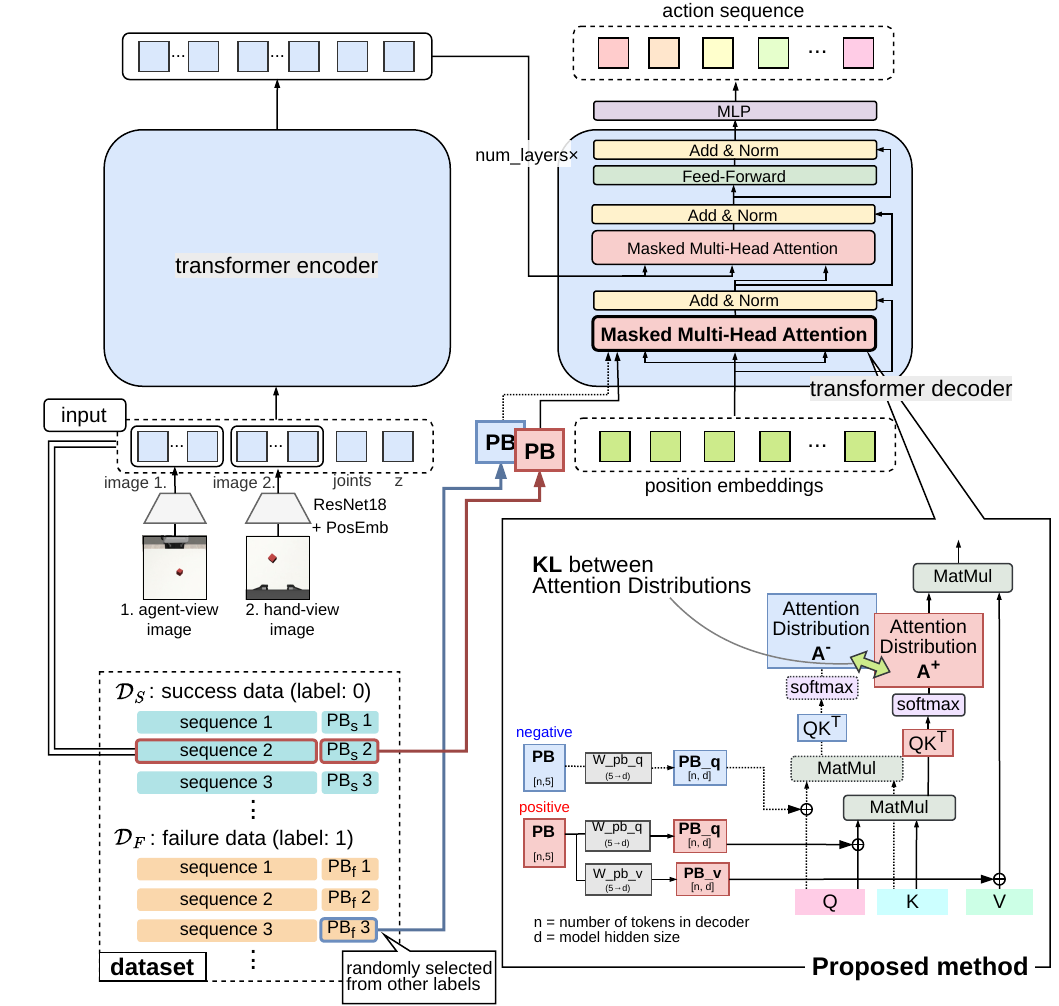}
  \caption{Overview of the proposed method. The boxed area indicates the proposed modules added to the baseline ACT.}
  \label{fig:model_fig}
\end{figure}

\subsection{Process 1: Initial Full-Dataset Training}
% \subsection{Process 1: Initial Full-Dataset Training with All Success/Failure Dataset}
\label{sec:pb_kl}
First, an imitation learning model is trained on all successful and failed demonstrations in the dataset.
We adopt Action Chunking with Transformers (ACT)~\cite{ACT} as the baseline imitation learning model.
ACT uses a transformer~\cite{Transformer} to predict a future action sequence $a_{t+1,\dots,t+H}$ from an observation $o_t$ consisting of multi-view camera images and joint angles.
In ACT, self-attention captures sequence dependencies and plays a key role in action generation by aggregating value vectors with query--key compatibility weights, thereby indicating which information is emphasized in the sequence.

In this study, we hypothesize that successful and failed trajectories place attention on different parts of the visual and state information during action generation, and that this discrepancy appears as a difference in their attention distributions.
Based on this hypothesis, we slightly modulate the representations in the decoder self-attention using data-specific latent variables, aiming to enable the formation of distinct attention patterns for successful and failed demonstrations.
To capture the distributional discrepancy between successful and failed data, we introduce PB, which is a learnable latent vector assigned to each training sample.
In our implementation, PB is defined as a five-dimensional vector for each token.
Here, $L$ denotes the length of the predicted action sequence.
The PB for each sample is represented as a sequence of vectors of size $L \times 5$, initialized to zero and updated through training.
While the weights of the main model are shared across all samples, the PBs are sample-specific.
Therefore, each PB can capture latent characteristics that are unique to its corresponding sample.

PB is incorporated into the self-attention mechanism of the transformer decoder in ACT, where it modifies the attention distribution by being added to the query and value representations.
Figure~\ref{fig:model_fig} illustrates the mechanism of the self-attention layer with PB.
At each layer, PB is projected into the attention space through independent, layer-specific linear mappings for query and value, and the resulting bias terms are scaled by a coefficient $\alpha_{\mathrm{pb}}$ before being added to the corresponding representations.
The resulting attention with PB is given by Eq.~(\ref{eq:attn_pb}).
\begin{equation}
\mathrm{Attention}(Q,K,V)
=
\mathrm{softmax}\!\left(
\frac{(Q+\alpha_{\mathrm{pb}}\mathrm{pb}_q)K^{\top}}{\sqrt{d_k}}
\right)
\,(V+\alpha_{\mathrm{pb}}\mathrm{pb}_v)
\label{eq:attn_pb}
\end{equation}
Here, $Q$, $K$, and $V$ denote the query, key, and value matrices, respectively, and $d_k$ is the dimensionality of the key (and query).
$\mathrm{pb}_q$ and $\mathrm{pb}_v$ denote the bias terms on the query side and the value side, respectively, obtained by projecting PB into the attention space through layer-specific linear mappings.

In the proposed method, we further introduce KL regularization, which encourages the separation of attention distributions so that PB can represent the difference between successful and failed demonstrations in terms of attention.
During training, for an input observation $o_t$, we compute the attention distribution obtained using the PB corresponding to the input itself (positive PB) as $A^{+}(o_t)$, and the attention distribution obtained using a PB randomly selected from data with a different label (negative PB) as $A^{-}(o_t)$.
% The KL divergence between these two distributions is then used as a measure of their distributional discrepancy (Fig.~\ref{fig:model_fig}).
The symmetrized KL divergence between these two distributions is then used as a measure of their distributional discrepancy (Fig.~\ref{fig:model_fig}).
In this study, to encourage learning in a direction that further separates the attention distributions of successful and failed demonstrations,
% we define a loss with a negative sign on the KL divergence as the KL regularization term $\mathcal{L}_{\mathrm{KL}}$.
we define a loss with a negative sign on the symmetrized KL divergence as the KL regularization term $\mathcal{L}_{\mathrm{KL}}$.
\begin{equation}
\mathcal{L}_{\mathrm{KL}}
=
-\,D_{\mathrm{sym}}\!\left(A^{+}(o_t), A^{-}(o_t)\right)
\label{eq:kl_loss}
\end{equation}
where
\begin{equation}
D_{\mathrm{sym}}(P,Q)
=
\frac{1}{2}
\left[
D_{\mathrm{KL}}(P\|Q)
+
D_{\mathrm{KL}}(Q\|P)
\right].
\label{eq:sym_kl}
\end{equation}
Here, $D_{\mathrm{KL}}(P\|Q)$ denotes the KL divergence between probability distributions $P$ and $Q$.
% \begin{equation}
% \mathcal{L}_{\mathrm{KL}}
% =
% -\,D_{\mathrm{KL}}\!\left(A^{+}(o_t)\,\|\,A^{-}(o_t)\right)
% \label{eq:kl_loss}
% \end{equation}
% Here, $D_{\mathrm{KL}}(P\|Q)$ denotes the KL divergence between probability distributions $P$ and $Q$.
The final training objective is defined by adding the KL regularization term, which is weighted by a coefficient $\lambda_{\mathrm{kl}}$, to the imitation learning loss of ACT, $\mathcal{L}_{\mathrm{ACT}}$, as follows.
\begin{equation}
\mathcal{L}
=
\mathcal{L}_{\mathrm{ACT}}
+
\lambda_{\mathrm{kl}}\,\mathcal{L}_{\mathrm{KL}}
\label{eq:total_loss}
\end{equation}
The model parameters and PB are jointly optimized to minimize $\mathcal{L}$.
In our experiments, $\lambda_{\mathrm{kl}}$ was set to a small value so that the KL term acted as a regularizer without dominating the imitation loss.
Consequently, PB is expected to form distinct attention patterns for successful and failed demonstrations, allowing their distributional difference to be expressed as a difference in attention.

% \subsection{Process 2: Failure Data Selection Based on the KL Metric and Re-Training}
\subsection{Process 2: KL-Based Failure Selection and Re-Training}
\label{sec:kl_select}

In the second process, after the initial full-dataset training, we quantify the difference in attention distributions for each failure sample by comparing the attention obtained with its own PB and that obtained with each successful PB. The model is then retrained using a reconstructed dataset based on this metric.
Specifically, for each failure sample $f \in \mathcal{D}_F$, we use its observation sequence $o_t^{f}$ as the input, and compare the attention distributions obtained when the PB corresponding to the failure sample $f$ is used as the positive PB and the PB corresponding to each successful sample $s \in \mathcal{D}_S$ is used as the negative PB.
% The metric for a failure sample $f$ is defined as Eq.~(\ref{eq:kf}) by, for each successful sample $s \in \mathcal{D}_S$, computing the KL divergence between these two attention distributions at each time step $t$, averaging it over time, and then summing the resulting values over all successful samples.
% The metric for a failure sample $f$ is defined as Eq.~(\ref{eq:kf}) by computing the KL divergence between these two attention distributions at each time step $t$ for each successful sample $s \in D_S$. Next, it is averaged over time, and the resulting values are then summed over all successful samples.
The metric for a failure sample $f$ is defined in Eq.~(\ref{eq:kf}). 
For each successful sample $s \in \mathcal{D}_S$, the symmetrized KL divergence between these two attention distributions is computed at each time step $t$ and averaged over time. 
The resulting values are then summed over all successful samples.
% \begin{equation}
% K(f)
% =
% \sum_{s \in \mathcal{D}_S}
% \left(
% \frac{1}{T}
% \sum_{t=1}^{T}
% D_{\mathrm{KL}}
% \!\left(
% A^{+}(o_t^{f};\mathrm{PB}_f)
% \,\|\,
% A^{-}(o_t^{f};\mathrm{PB}_s)
% \right)
% \right)
% \label{eq:kf}
% \end{equation}
\begin{equation}
K(f)
=
\sum_{s \in \mathcal{D}_S}
\left(
\frac{1}{T}
\sum_{t=1}^{T}
D_{\mathrm{sym}}
\!\left(
A^{+}(o_t^{f};\mathrm{PB}_f),
A^{-}(o_t^{f};\mathrm{PB}_s)
\right)
\right)
\label{eq:kf}
\end{equation}
Here, $A^{+}(o_t^{f}; \mathrm{PB}_f)$ denotes the attention distribution obtained by applying the positive PB, $\mathrm{PB}_f$, to the observation $o_t^{f}$ of the failure sample $f$, and $A^{-}(o_t^{f}; \mathrm{PB}_s)$ denotes the attention distribution obtained by applying the negative PB, $\mathrm{PB}_s$, to the same observation $o_t^{f}$.
In addition, $T$ denotes the number of time steps over which the attention distributions are compared for the failure sample $f$.
Furthermore, $K(f)$ is computed from multiple training runs with different random seeds, and its average value, $\bar{K}(f)$, is used as the final metric.

Based on this metric $\bar{K}(f)$, the failure samples are ranked in ascending order; those with smaller values are selected and combined with the successful demonstrations to reconstruct the training dataset $\{\mathcal{D}_S, \mathcal{D}_F^{\mathrm{low}}\}$, on which the model is retrained.
A small value of $\bar{K}(f)$ for a failure sample indicates that the set of successful demonstrations contains examples whose environmental states and trajectories are relatively similar to those of the failure sample.
% Such failure samples may include cases that share similar environmental conditions and trajectories with successful demonstrations, but result in failure owing to only slight differences.
Such failures may correspond to near-success cases that fail owing to only slight differences.
% Based on the hypothesis that it is easier to learn the essential factors that determine success or failure by distinguishing between closely related successful and failed examples, rather than between largely dissimilar ones, this study selects failure samples with small KL-based metrics with respect to the successful data for training.
% We hypothesize that it is easier to learn the essential factors that determine success or failure by distinguishing between closely related successful and failed examples, rather than between largely dissimilar ones.
We hypothesize that the essential factors determining success or failure are easier to learn by distinguishing between closely related successful and failed examples than between largely dissimilar ones.
Based on this hypothesis, we select failure samples with small values of this KL-based metric with respect to successful demonstrations for training.

\subsection{PB Selection at Inference Time}
\label{sec:pb_select}

% At inference time, the nearest successful sample from the training data is selected based on the embedding of the initial overhead observation image, and its corresponding PB is used for action generation.
At inference time, we select the nearest successful training sample based on the embedding of the initial overhead observation image and use its corresponding PB for action generation.
Specifically, after training, the initial overhead image of each successful training sample is embedded using ResNet18~\cite{resnet} and stored.
During inference, the input initial image is embedded in the same manner, and the nearest successful sample is identified using cosine similarity to the stored embeddings.

In this method, PB is not directly estimated from the initial image.
Instead, the initial image is used as a cue to retrieve a similar successful episode, and the associated PB is used.
Because PB may contain not only initial-state information but also trajectory-specific factors, this approach should be regarded as approximate retrieval-based PB selection rather than exact PB estimation.
Nevertheless, successful episodes with similar initial states are likely to share the approach and grasping conditions required for success.
Moreover, through KL regularization, successful PBs are learned to represent success-oriented attention modes distinct from failed PBs.
Therefore, using the PB from a similar successful episode can bias action generation toward successful behavior during inference.

%% file: sections/4_experiments.tex
\section{Experiments}
\subsection{Simulation Environment and Training Setup}
To evaluate the proposed method, we conducted experiments using the Lift task, which is a cube-lifting task in a simulation environment.
We used robosuite~\cite{robosuite}, which is built on MuJoCo~\cite{mujoco}, and employed a seven-degree-of-freedom (7-DoF) Panda robot arm.
A red cube with an edge length of approximately 4\,cm was placed on a table, and its initial position was randomly sampled within 
$x \in [-0.1, 0.1]\,$m and $y \in [-0.2, 0.2]\,$m in the table coordinate frame.
A trial was considered successful when the cube center was lifted more than 4\,cm above the table.

\begin{figure}[htbp]
    \centering
    \includegraphics[width=\columnwidth]{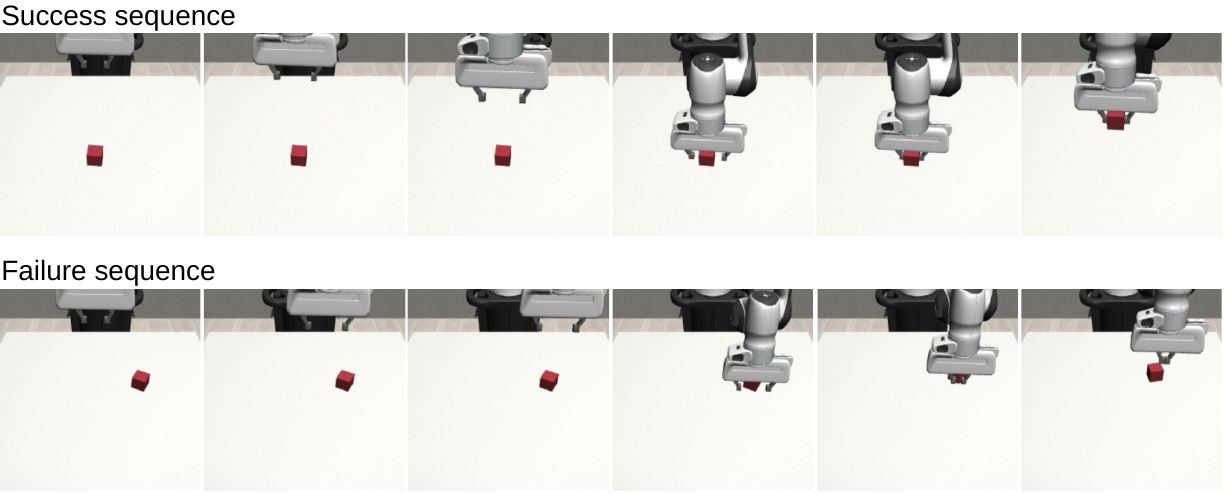}
    \caption{Examples of successful (top) and failed (bottom) sequences in the Lift task.}
    \label{fig:lift_s_f}
\end{figure}

We collected 100 teleoperated trajectories using a three-dimensional (3D) mouse, comprising 50 successful and 50 failed trajectories. 
These demonstrations form the full dataset $\{\mathcal{D}_S, \mathcal{D}_F\}$, where $\mathcal{D}_S$ and $\mathcal{D}_F$ denote the successful and failed subsets, respectively.
Figure~\ref{fig:lift_s_f} shows representative examples of successful and failed trajectories.
The model input consisted of two-view RGB images, an overhead image and a wrist-camera image, each with a resolution of $224 \times 224 \times 3$, and a 14-dimensional robot state composed of seven-dimensional joint angles, a six-dimensional end-effector pose, and a one-dimensional gripper opening.
% Each model was trained for 10,000 epochs with a batch size of 10 and a chunk size of 100.
% The PB shape was therefore $[100, 5]$.
% AdamW was used for optimization.
% Each model was trained using five seeds, from 0 to 4.
% Each model was trained for 10,000 epochs using AdamW with a batch size of 10 and a chunk size of 100.
% The PB shape was therefore $[100, 5]$.
% Training was performed using five random seeds, from 0 to 4.
Each model was trained for 10,000 epochs using AdamW with a batch size of 10, a chunk size of 100, and five random seeds from 0 to 4.
The PB shape was therefore $[100, 5]$.

\subsection{Evaluation Protocol}
In each experiment, we used the task success rate as the performance metric.
We evaluated the models trained with five random seeds and performed 100 rollouts per seed.
The final performance was reported as the average success rate over five seeds.
In this study, we conducted the following two experiments.

% \textbf{Experiment 1: Effectiveness of Proposed Method}
\subsubsection{Experiment 1: Effectiveness of Proposed Method}

We used ACT trained only on 50 successful demonstrations $\mathcal{D}_S$ as the baseline.
We then compared it with ACT trained on the full demonstration dataset $\{\mathcal{D}_S, \mathcal{D}_F\}$, consisting of 50 successful and 50 failed demonstrations, as a conventional approach.
In addition, we evaluated the proposed method trained on $\mathcal{D}_S$ and on $\{\mathcal{D}_S, \mathcal{D}_F\}$.
Through these comparisons, we examined the effectiveness of incorporating failure data using the proposed method.
 
% \textbf{Experiment 2: Analysis of Failure Data Selection Strategies}
\subsubsection{Experiment 2: Analysis of Failure Data Selection Strategies}

To evaluate the effect of failure data selection on task success rates, we compared random selection with our KL-based selection method.
% To evaluate how failure-data selection affects task success rates, we compared random selection with selection based on the KL-based metric.
In random selection, the 50 failed demonstrations $\mathcal{D}_F$ were randomly divided into five disjoint subsets of 10 demonstrations each.
Each subset was combined with the 50 successful demonstrations $\mathcal{D}_S$ to construct five datasets under the ``50 successes + 10 failures'' setting.
ACT and the proposed method were trained on these datasets to evaluate the effect of randomly selected failure data.
% For selection based on the KL-based metric, we sorted the failed demonstrations in ascending order according to the value of $\bar{K}(f)$.
For KL-based selection, failed demonstrations were sorted in ascending order of $\bar{K}(f)$, and three subsets were selected:
(i) $\mathcal{D}_F^{\mathrm{low}}$, the 10 demonstrations with the smallest values;
(ii) $\mathcal{D}_F^{\mathrm{mid}}$, 10 demonstrations around the median values; and
(iii) $\mathcal{D}_F^{\mathrm{high}}$, the 10 demonstrations with the largest values.
Each subset was combined with $\mathcal{D}_S$ to construct three reconstructed datasets, which were used to train both ACT and the proposed method.
These experiments compare random and KL-based selection and analyze the effect of using failure data from different KL-value ranges.

%% file: sections/5_result.tex
\section{Results and Discussion}

This section presents the experimental results obtained on the Lift task to evaluate the effectiveness of incorporating failure data when training with the proposed method. 
We also investigate whether failure data that effectively contribute to learning can be selected using a post-training analysis metric. 
All results are reported as the mean and standard deviation of the success rate over five random seeds.

\subsection{Experiment 1: Effectiveness of Proposed Method}

\begin{table}[H]
    \centering
    \caption{Average task success rates for each model in Experiment 1.} 
    \label{tab:result1}
    \begin{tabular}{lcc}
    \hline
        Model & Dataset & Success rate [\%] \\ \hline \hline
        ACT & $\mathcal{D}_S$ & $71.0 \pm 3.39$ \\ \hline 
        ACT & $\{\mathcal{D}_S,\mathcal{D}_F\}$ & $71.6 \pm 5.22$ \\ \hline
        Proposed method\textsuperscript{*}& $\mathcal{D}_S$ & $74.2 \pm 3.63$ \\ \hline 
        Proposed method & $\{\mathcal{D}_S,\mathcal{D}_F\}$ & $\mathbf{77.8 \pm 2.59}$ \\ \hline
    \end{tabular}

    \vspace{1mm}
    \scriptsize
    \textsuperscript{*}The KL term is inactive because no failure data are used.
\end{table}

Table~\ref{tab:result1} presents the results of Experiment 1.
% First, we compare ACT trained only on $\mathcal{D}_S$ as the baseline with ACT trained on $\{\mathcal{D}_S,\mathcal{D}_F\}$.
First, we use ACT trained only on $\mathcal{D}_S$ as the baseline and compare it with ACT trained on $\{\mathcal{D}_S,\mathcal{D}_F\}$.
The average success rates were 71.0\% and 71.6\%, respectively, showing no substantial difference.
In addition, the standard deviation of the success rate increased when failure data were added.
These results indicate that simply adding failure data does not necessarily improve task performance.
When the proposed method was trained only on 50 successful demonstrations, the attention distributions of successful and failed demonstrations could not be compared.
Therefore, the KL regularization term was not effectively applied, and this condition primarily evaluates the effect of introducing PB.
Even under this condition, the success rate reached 74.2\%, outperforming ACT.
This result indicates that introducing PB itself does not degrade performance and may contribute to a modest improvement.
% Even under this condition, the success rate reached 74.2\%, outperforming ACT, indicating that the introduction of PB itself did not degrade performance, and might result in improvements.
Furthermore, the proposed method trained on $\{\mathcal{D}_S,\mathcal{D}_F\}$ achieved the highest success rate of 77.8\%.
By jointly using PB and KL regularization with failure data, the proposed method achieved a performance improvement that was not obtained by simply adding failure data to ACT.
% These results demonstrate the effectiveness of the proposed method for learning that makes effective use of failure data.
% These results demonstrate the effectiveness of the proposed method for learning by effectively utilizing failure data.
These results demonstrate that the proposed method effectively leverages failure data for learning.

\begin{figure}[tb]
    \begin{center}
    \includegraphics[width=\columnwidth]{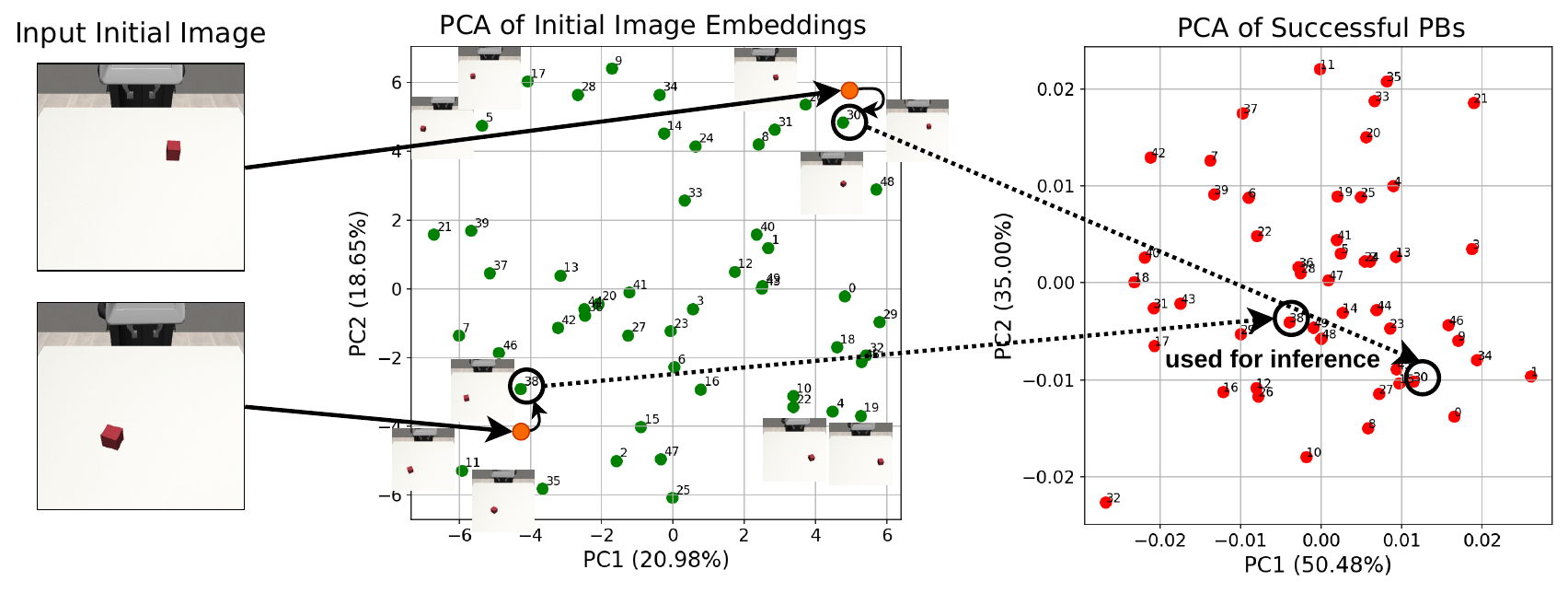}
  % \caption{PB selection during inference. The initial observation image is embedded, and the PB of the nearest successful training sample is used for action generation. Several points in the image embedding PCA space are shown with their corresponding initial observation images.}
  \caption{PB selection during inference using nearest-neighbor retrieval in the initial observation embedding space. Several points are shown with their corresponding initial observation images.
  }
  \label{fig:pb_search}
    \end{center}
\end{figure}

In addition, Fig.~\ref{fig:pb_search} illustrates the process of retrieving a similar successful training sample from the initial observation image during inference and using the corresponding PB for action generation.
Principal component analysis (PCA) was used to project each feature representation, including the image features and PBs, into a low-dimensional space.
Comparing the images mapped in the image feature space at the center of the figure with the initial observation image at inference time, the method selects a training sample with a similar object position.
The PB corresponding to this similar successful sample is then used for action generation.
% This PB selection strategy is likely one reason why the proposed method improved the success rate even when no failure data were used.
This PB selection strategy might have contributed to the improved success rate using the proposed method, even when no failure data were used.

In the proposed method, PB represents trajectory-specific characteristics, while KL regularization encourages the formation of different attention distributions for successful and failed demonstrations.
Therefore, even under the condition that includes failure data, the proposed method not only increases the diversity of the training distribution but also enables learning while distinguishing the differences between success and failure in the attention mechanism.
This property is considered to have contributed to the performance improvement.
In addition, because the proposed method selects a success-side PB during inference for action generation, the stable formation of success-side attention patterns during training can be interpreted as leading to the final improvement in the success rate.
These results indicate that whereas conventional ACT does not achieve sufficient improvement by simply adding failure data, the proposed method can effectively utilize failure data for learning.

\subsection{Experiment 2: Analysis of Failure Data Selection Strategies}
\begin{figure}[tb]
    \begin{center}
    \includegraphics[width=\columnwidth]{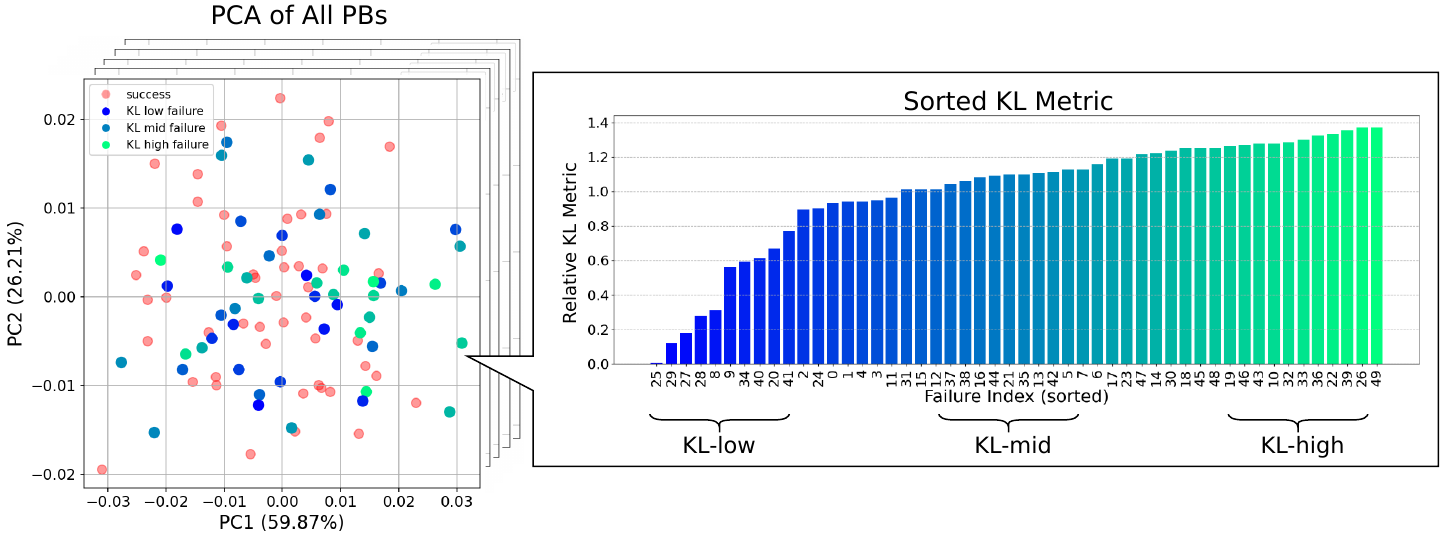}
  % \caption{Overview of failure data selection based on the KL metric.
  %   The figure shows the PCA projection of PBs and the KL metrics of failure samples sorted in ascending order and shown relative to the minimum value.
  %   The metric aggregates the discrepancy between attention distributions obtained with each fixed failure PB and all successful PBs.}
    \caption{
    Overview of failure data selection based on the KL metric.
    The left panel shows the PCA projection of all PBs, with red for successful PBs and blue-to-green for failure PBs according to their KL metric values.
    The blue-to-green gradient indicates increasing KL metric values and corresponds to the colors in the right panel, where failure samples are sorted by the metric averaged over five training runs relative to the minimum value.}
  \label{fig:pb_kl_fig}
    \end{center}
\end{figure}

In Experiment 2, we evaluate failure samples using the model previously trained on the full demonstration dataset $\{\mathcal{D}_S,\mathcal{D}_F\}$ in Experiment 1, then reconstruct the dataset and retrain the model.
% We evaluate failure data using the model trained on the full demonstration dataset $\{\mathcal{D}_S,\mathcal{D}_F\}$ in Experiment 1, reconstruct the dataset, and retrain the model.
Figure~\ref{fig:pb_kl_fig} presents the KL metrics of failure samples sorted in ascending order and shown relative to the minimum value.
Based on this ranking, we divide the failure data into three groups, $\mathcal{D}_F^{\mathrm{low}}$, $\mathcal{D}_F^{\mathrm{mid}}$, and $\mathcal{D}_F^{\mathrm{high}}$, and select the failure data used for training.

Table~\ref{tab:result2} lists the training results for each dataset.
For the five datasets constructed by random failure selection, ACT achieved an average success rate of 72.8\%, whereas the proposed method achieved 75.8\%.
This indicates that the proposed method can exploit even a small number of failure samples, although random selection may cause performance variation depending on the selected failures.
% \begin{table}[h]
%     \centering
%     \caption{Average task success rates [\%] for each dataset and model in Experiment 2.}
%     \label{tab:result2}
%     \begin{tabular}{lcc}
%         Data & ACT & Proposed method \\ \hline \hline
%         $\mathcal{D}_S$ & $71.0 \pm 3.39$ & - \\ \hline
%         $\{\mathcal{D}_S,\mathcal{D}_F^{\mathrm{rand}}\}$\textsuperscript{*} & $72.8 \pm 5.55$ & $75.8 \pm 5.40$ \\ \hline
%         $\{\mathcal{D}_S,\mathcal{D}_F^{\mathrm{low}}\}$ & $74.2 \pm 7.40$ & $\mathbf{79.4 \pm 4.56}$ \\ \hline
%         $\{\mathcal{D}_S,\mathcal{D}_F^{\mathrm{mid}}\}$ & $76.0 \pm 5.74$ & $78.6 \pm 2.07$ \\ \hline
%         $\{\mathcal{D}_S,\mathcal{D}_F^{\mathrm{high}}\}$& $70.4 \pm 7.44$ & $71.8 \pm 3.35$ \\ \hline
%     \end{tabular}
    
%     \vspace{1mm}
%     \scriptsize
%     \textsuperscript{*}This result is averaged over five randomly selected datasets.
% \end{table}

\begin{table}[h]
    \centering
    \caption{Average task success rates for each dataset and model in Experiment 2.}
    \label{tab:result2}
    \begin{tabular}{ccc}
    \hline
    \multirow{2}{*}{Dataset} & \multicolumn{2}{c}{Success rate [\%]} \\ \cline{2-3}
                             & ACT & Proposed method \\ \hline \hline
    $\mathcal{D}_S$ & $71.0 \pm 3.39$ & - \\ \hline
    $\{\mathcal{D}_S,\mathcal{D}_F^{\mathrm{rand}}\}$\textsuperscript{*} 
        & $72.8 \pm 5.55$ & $75.8 \pm 5.40$ \\ \hline
    $\{\mathcal{D}_S,\mathcal{D}_F^{\mathrm{low}}\}$ 
        & $74.2 \pm 7.40$ & $\mathbf{79.4 \pm 4.56}$ \\ \hline
    $\{\mathcal{D}_S,\mathcal{D}_F^{\mathrm{mid}}\}$ 
        & $76.0 \pm 5.74$ & $78.6 \pm 2.07$ \\ \hline
    $\{\mathcal{D}_S,\mathcal{D}_F^{\mathrm{high}}\}$ 
        & $70.4 \pm 7.44$ & $71.8 \pm 3.35$ \\ \hline
\end{tabular}
    
    \vspace{1mm}
    \scriptsize
    \textsuperscript{*}This result is averaged over five randomly selected datasets.
\end{table}

Under KL-based selection, the proposed method achieved 79.4\%, 78.6\%, and 71.8\% under the $\mathcal{D}_F^{\mathrm{low}}$, $\mathcal{D}_F^{\mathrm{mid}}$, and $\mathcal{D}_F^{\mathrm{high}}$ conditions, respectively.
Among all settings, the $\mathcal{D}_F^{\mathrm{low}}$ condition gave the highest success rate, and the success rate tended to decrease as the KL metric increased.
In particular, under $\mathcal{D}_F^{\mathrm{high}}$, both ACT and the proposed method performed below the random average and close to the baseline condition.
The KL metric $\bar{K}(f)$ represents the discrepancy between the attention distribution using the PB of each failure sample $f$ and those using successful PBs.
Although the proposed method separates success and failure attention distributions overall, the degree of separation varies among failure samples.
This suggests that failures with smaller discrepancies from successful samples are more useful for training.
% Failure samples with small KL metrics are likely to have relatively similar successful examples in terms of environment states or trajectories.
% Thus, they are likely to share much of their structure with successful demonstrations, but fail owing to subtle differences.
Failure samples with small KL metrics are likely to share environmental states or trajectories with successful demonstrations but fail owing to subtle differences.
Training with such failures is considered to help the model learn not only a coarse success/failure distinction but also fine-grained factors that differentiate successful and failed behaviors.
In contrast, failure samples with large KL metrics may share less structure with successes and mainly reflect differences in environmental conditions or overall trajectories, making them less useful for improving successful action generation.
Comparing the random average with $\mathcal{D}_F^{\mathrm{low}}$, the success rate improved from 72.8\% to 74.2\% for ACT and from 75.8\% to 79.4\% for the proposed method.
These results indicate that selecting failures that complement successful demonstrations based on the KL metric can further improve performance, rather than simply adding failure data.
% These results indicate that selecting failures that complement successful demonstrations is more important than simply adding more failure data.

\begin{figure}[tb]
    \begin{center}
    \includegraphics[width=\columnwidth]{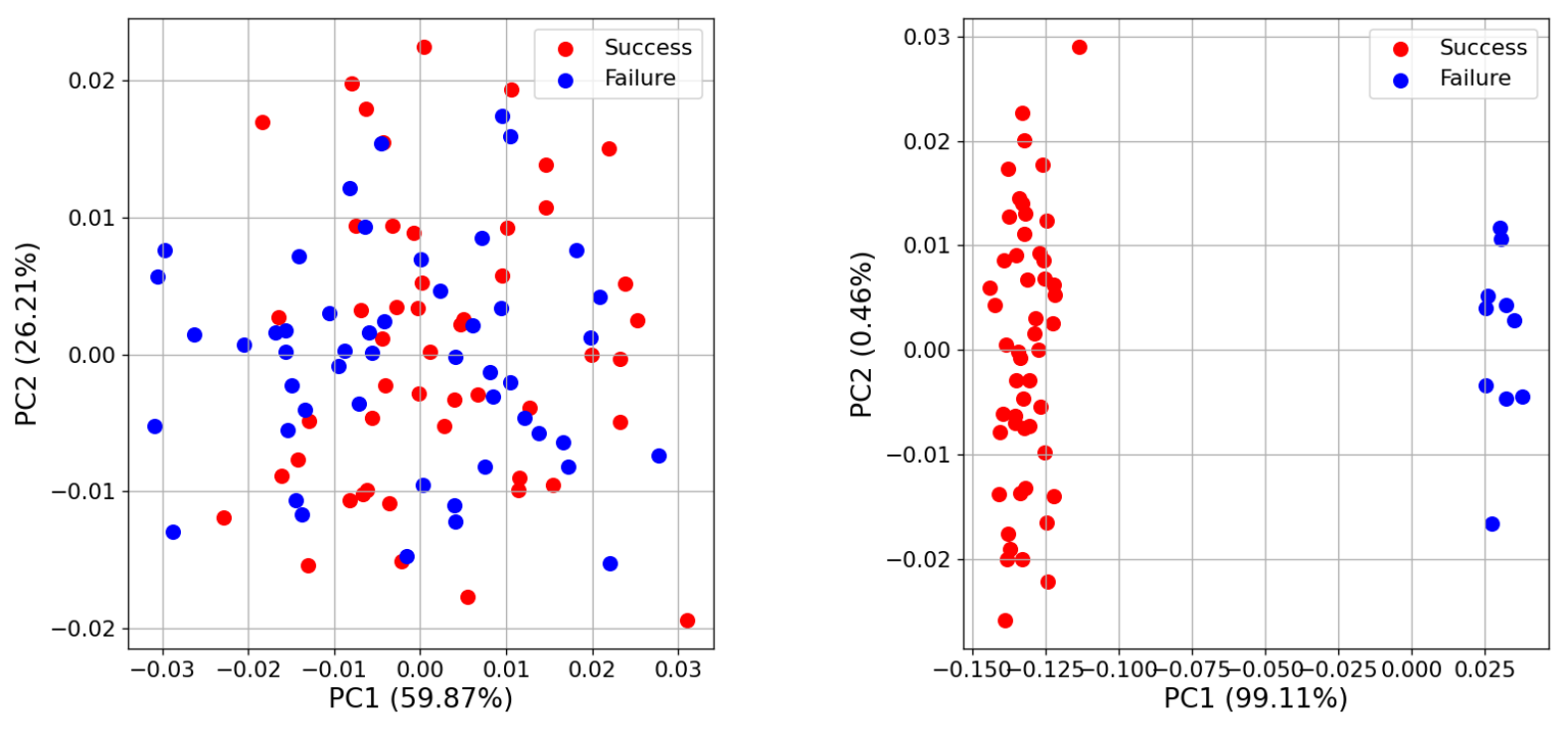}
  \caption{PCA visualization of PBs learned by the proposed method under $\{\mathcal{D}_S,\mathcal{D}_F\}$ (left) and $\{\mathcal{D}_S,\mathcal{D}_F^{\mathrm{low}}\}$ (right). Red and blue points denote PBs of successful and failed demonstrations, respectively.}
  \label{fig:ex1_ex2_pb}
    \end{center}
\end{figure}

Figure~\ref{fig:ex1_ex2_pb} shows PCA visualizations of PBs learned by the proposed method under the $\{\mathcal{D}_S,\mathcal{D}_F\}$ condition in Experiment 1 (left) and the $\{\mathcal{D}_S,\mathcal{D}_F^{\mathrm{low}}\}$ condition in Experiment 2 (right).
Successful and failed PBs largely overlap in the left plot, whereas they are more clearly separated in the right plot.
This indicates that selecting a small number of failures can make success--failure differences more apparent in the PB space.
% This indicates that the selection of a small number of failures can make success/failure differences more apparent in the PB space.
When many failures are uniformly added, diverse failures that share little structure with successes may also be included, making the success--failure differences in the PB space more diffuse.
In contrast, selecting a small number of useful failures may make the learned differences clearer.
% However, PCA alone cannot determine the effectiveness of PB learning.
However, the effectiveness of PB learning cannot be determined using only PCA.
Because the proposed method still outperformed ACT under $\{\mathcal{D}_S,\mathcal{D}_F\}$, unclear PCA separation does not necessarily imply ineffective PBs.

% These results indicate that selecting failures that are useful for successful demonstrations is more important than simply adding more failure data.
% The PB-space visualization can be considered supporting evidence for this interpretation.
% The results also indicate that the KL metric can serve as an effective criterion for selecting useful failure data.

These results indicate that selecting failures useful for successful demonstrations is more important than simply adding more failure data, and that the KL metric can serve as an effective selection criterion.

%% file: sections/6_conclusion.tex
\section{Conclusion}
In this study, we proposed a method for directly exploiting failure data accumulated during data collection in imitation learning and selecting useful failure samples using an attention-based metric.
We introduced PB, a learnable parameter assigned to each demonstration, into a transformer encoder--decoder imitation learning model.
By adding KL regularization between the attention distributions of successful and failed demonstrations in the transformer decoder multi-head attention, PBs were encouraged to capture success--failure differences.
Experiments on the simulated Lift task showed that incorporating failure data with PB and KL regularization improved task success rates, and that the KL metric helped select failure data that were more effective for training.
% These results indicate that the proposed method is effective for utilizing failure data, and the KL metric is a promising criterion for selecting useful failure data.
In future work, we plan to extend our method to real-robot environments and other manipulation tasks.